\documentclass{article} 
\usepackage{iclr2027_conference,times}

\usepackage{amsmath,amsfonts,bm}

\def\eqref#1{equation~\ref{#1}}

\def\1{\bm{1}}

\DeclareMathAlphabet{\mathsfit}{\encodingdefault}{\sfdefault}{m}{sl}
\SetMathAlphabet{\mathsfit}{bold}{\encodingdefault}{\sfdefault}{bx}{n}

\usepackage[most]{tcolorbox}
\usepackage{xcolor}

\newtcolorbox{theorybox}{
  enhanced,
  breakable,
  colback=blue!3,
  colframe=blue!30!black,
  boxrule=0.5pt,
  arc=2pt,
  left=5pt,
  right=5pt,
  top=4pt,
  bottom=4pt,
  before skip=3pt,
  after skip=5pt
}
\AtBeginDocument{
  \setlength{\abovedisplayskip}{5pt plus 2pt minus 2pt}
  \setlength{\belowdisplayskip}{5pt plus 2pt minus 2pt}
  \setlength{\abovedisplayshortskip}{3pt plus 2pt minus 2pt}
  \setlength{\belowdisplayshortskip}{3pt plus 2pt minus 2pt}
}

\usepackage{hyperref}
\usepackage{url}
\usepackage{booktabs}
\usepackage{multirow}
\usepackage{amsmath}
\usepackage{graphicx}
\usepackage{amssymb}
\usepackage{amsthm}

\usepackage{longtable}

\usepackage{multirow}
\usepackage[table]{xcolor}
\usepackage{graphicx}

\definecolor{skillgreen}{RGB}{218,242,218}
\definecolor{taskpink}{RGB}{252,240,240}
\definecolor{taskorange}{RGB}{252,246,236}
\definecolor{taskblue}{RGB}{241,242,252}

\usepackage{color}

\title{SkillMem: Compact External Skill Memory for Latent Multi-Agent Reasoning}

\author{%
\textbf{Yuxuan Zhu}\textsuperscript{1},
\textbf{Mohammad Mohammadi Amiri}\textsuperscript{1},
\textbf{Pin-Yu Chen}\textsuperscript{2},
\textbf{Keerthiram Murugesan}\textsuperscript{2}
\\[0.4em]
{\small
\textsuperscript{1}Rensselaer Polytechnic Institute
\quad
\textsuperscript{2}IBM Research
}\\[0.2em]
{\small\texttt{\{zhuy27,mamiri\}@rpi.edu}},
{\small\texttt{\{pin-yu.chen,keerthiram.murugesan\}@ibm.com}}
}

\iclrfinalcopy

\begin{document}

\maketitle

\lhead{}

\begin{abstract}

Latent multi-agent systems reduce inference cost by keeping intermediate
reasoning and inter-agent communication in continuous representations.\
Reusable skills, however, are commonly supplied as text or stored as
layer-wise Key-Value states or skill-specific model parameters, introducing repeated
processing or substantial per-skill storage.\ We introduce
\textbf{SkillMem}, which represents each reusable skill as a small external
latent memory and makes it directly available during latent multi-agent
reasoning.\ A role-conditioned cross-attention reader allows the same memory
to guide different agents according to their roles and current hidden states.\
We train the skill memories and shared reader through behavioral distillation,
matching the change in token predictions induced by a textual skill while
keeping the backbone frozen.\ For our experiments, we obtain textual skills
from paired successful and failed multi-agent trajectories, although existing
skills can also be compiled directly.\ Across six benchmarks and Qwen3 models ranging from 4B to 14B parameters,
SkillMem provides strong gains for smaller backbones while preserving
efficient latent multi-agent inference across model scales.\ We use the same
$8\times256$ memory configuration across all evaluated model scales, requiring
9.6\,KB of skill-specific storage per skill, including its retrieval
embedding.\ For a library of 100 skills, SkillMem reduces total storage by more than $38\times$ compared with alternative skill representations and
achieves up to a $4.23\times$ end-to-end inference speedup.\ These
results show that compact procedural memories can be reused without
sacrificing the efficiency of latent multi-agent reasoning.

\end{abstract}

\section{Introduction}
\label{sec:introduction}

Large language model (LLM) based multi-agent systems (MAS) assign different
roles to agents, allowing them to plan, critique, and refine a solution
across multiple reasoning steps.\ This division of work can improve performance over a single model
response, yet role specialization alone does not prevent recurring errors.
Some of these errors can be addressed by procedures such as checking assumptions or verifying intermediate results.\ Skills~\citep{zhang2025agentskills} capture such reusable procedures
from previous experience and make them available for
similar problems.\ In text-based MAS, 
agents express their reasoning and
exchange intermediate results in natural language.\ Latent reasoning and latent MAS replace this text-based reasoning and communication with continuous representations, reducing the need to generate and process intermediate text~\citep{hao2024coconut,shen2025codi,
zou2026latentmas}.\ A textual skill is typically inserted into an agent's prompt,
so the model must prefill the skill tokens each time the skill is used.
Across repeated uses, this extra prefilling can erode the inference
speed gained from latent reasoning and communication. This raises a question: \textit{how should agents
reuse skills efficiently when their reasoning and communication remain in latent
space?}\ 

\begin{figure*}[t]
    \centering
    \includegraphics[width=0.9\textwidth]{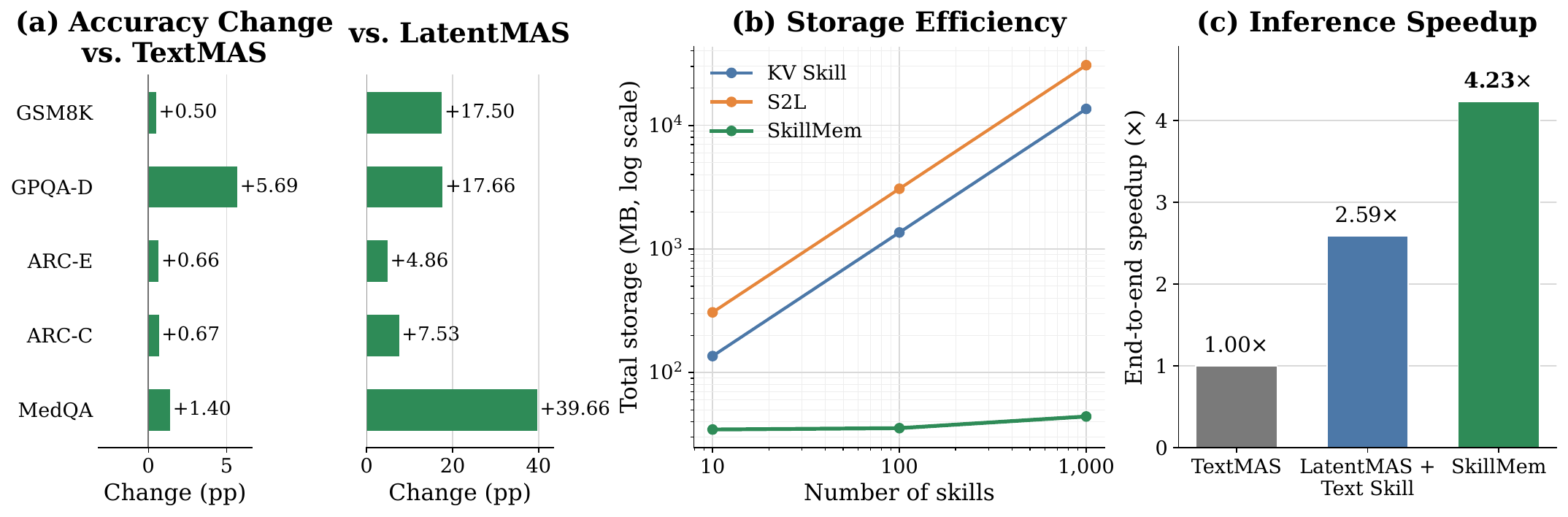}
    \caption{
    \textbf{Accuracy and efficiency comparison.}
    (a) Accuracy changes of SkillMem relative to TextMAS and LatentMAS
 using Qwen3-4B.
    (b) Total storage required by different skill representations as
    the skill library grows.
    (c) End-to-end inference speedup relative to TextMAS.
    }
    \vspace{-15pt}
    \label{fig:main_summary}
\end{figure*}

To avoid repeatedly processing skill text, we can store skills in other forms.\ For example, a cached key–value (KV) prefix avoids recomputing a skill prompt’s activations, but its storage grows with the prefix length and the number of layers~\citep{jin2026agentprimitives}.
A skill-specific LoRA adapter stores parameter updates for the layers it modifies~\citep{zhang2026skilltolora}.\ Although each adapter is smaller than the full model, storing one for every skill can still require substantial storage as the library grows.\ These trade-offs motivate a
compact but efficient, retrievable skill representation that agents
can use directly during latent reasoning.

We introduce \textbf{SkillMem}, a framework that stores each reusable
skill as a small set of learned latent memory vectors.\ The memory is external
to the frozen backbone and can be retrieved independently of other skills.
Given a task query, SkillMem selects a relevant skill and makes its
memory available throughout latent multi-agent reasoning.\ Agents
continue to exchange latent states, while the retrieved skill provides
procedural guidance without adding its textual description to each
agent's prompt.\ This separates the choice of a skill from the way an
agent uses it.

Sharing a skill across agents requires more than giving every role the
same memory vectors.\ A planner and a critic may use the same procedure
at different stages and for different purposes.\ SkillMem therefore
uses a role-conditioned \textbf{reader} to access the retrieved \textbf{memory embedding} at
selected layers of the frozen backbone.\ The memory is stored once,
while the reader uses the current agent's role to produce a
role-dependent update to its hidden states.\ This allows agents to
share procedural knowledge without requiring a separate skill
representation for each role.

The remaining challenge is how to learn a useful skill memory.\ Rather than training the memory to
reconstruct the skill description, we train it to preserve the
skill's effect on model behavior.\ We first generate a response using the textual skill.\ We then
measure how much the skill changes the probability of each token
in that same response.\ Behavioral distillation trains the memory
and shared reader to reproduce these changes while keeping the
backbone frozen.\ In our experiments, the textual skills come
from paired successful and failed multi-agent answer trajectories for the
same task.\ Comparing these trajectories helps identify reusable
procedures.\ The compilation step takes a textual skill as input, so it can use
either a skill discovered from trajectories or an existing skill
that has already proved useful.

Across six reasoning, question-answering, and code-generation benchmarks, SkillMem improves on the latent MAS baseline.\ On Qwen3-4B,\ accuracy rises from 17.97\% to 35.63\% on GPQA-Diamond and from 72.40\% to 89.90\% on GSM8K.\ Each skill requires only 9.6\,KB of skill-specific storage.\ SkillMem achieves a 4.23$\times$ end-to-end speedup over TextMAS and a 1.63$\times$ speedup over LatentMAS with text skills.\ With a library of 100 skills, its total storage, including the shared reader, is 38.3$\times$ smaller than KV Skill and 86.6$\times$ smaller than LoRA Skill.\ These results show that SkillMem can provide reusable skills for latent multi-agent reasoning, as summarized in Figure~\ref{fig:main_summary}.

\section{Related Work}
\label{sec:related}

\noindent\textbf{Latent multi-agent reasoning and memory.}
LatentMAS~\citep{zou2026latentmas} enables agents to reason and
communicate through continuous states rather than exchanging
natural-language messages.\ KVComm~\citep{ye2025kvcomm} and
RecursiveMAS~\citep{zou2026recursivemas} further study efficient
communication and computation in multi-agent systems.\ Building on
latent collaboration, LatentMem~\citep{fu2026latentmem} constructs
agent-specific memories from retrieved interaction experience, while
Agent Primitives~\citep{jin2026agentprimitives} represents predefined
multi-agent operations with KV-based latent states.\ SkillMem addresses
a different unit of reuse: a task-level procedural skill stored as
a persistent memory that can be retrieved and shared across agent
roles.

\noindent\textbf{Skill representations and behavioral distillation.}
Textual skills can be applied through prompting, but their instructions
must be processed whenever they are inserted into an agent's context.
S2L~\citep{zhang2026skilltolora} and
LatentSkill~\citep{yu2026latentskill} compile textual skills into
LoRA-based model adaptations.\ SkillSmith~\citep{dery2026skillsmith}
studies the composition of learned KV-prefix skills and textual
knowledge.\ SEED~\citep{wu2026seed} distills skill-induced behavioral
changes into the agent policy.\ SkillMem likewise learns from the
behavioral effect of a textual skill, but stores that effect in an
independently retrievable external memory.\ A reader shared across
skills and roles accesses the memory while the backbone remains
frozen.\ We compare against S2L as a parameter-based skill
representation in our experiments.

\noindent\textbf{Continuous representations and context compression.}
Prompt and prefix tuning~\citep{lester2021power,li2021prefix}
learn continuous representations that condition a frozen model.
Gisting and AutoCompressor~\citep{mu2023learning,
chevalier2023adapting} compress information from a textual context
into continuous representations.\ These methods motivate compact
continuous conditioning, while SkillMem learns a separate memory
for each reusable procedure and accesses it through a shared,
role-conditioned reader during latent multi-agent reasoning.

\noindent\textbf{Experience-based skill learning and routing.}
Voyager~\citep{wang2024voyager} and
ExpeL~\citep{zhao2023expel} reuse knowledge from previous agent
trajectories.\ SkillRL~\citep{xia2026skillrl} uses successful and
failed single-agent trajectories to build a hierarchical skill bank
and jointly updates the skills and agent policy through reinforcement
learning.\ We derive textual skills by comparing successful and
failed multi-agent trajectories, retaining the
roles involved in each reasoning stage.\ SkillRouter~\citep{
zheng2026skillrouter} studies skill selection at scale.\ Our
retriever selects a skill for the task, while SkillMem focuses on
how the selected skill is stored and used across agents.

\section{Preliminaries: Latent Multi-Agent Reasoning}
\label{sec:prelim-latent}

Let $f_\theta$ denote a decoder-only Transformer with $L$ layers.
During autoregressive generation, the model retains previously
computed KV states to avoid recomputing earlier
positions.\ Latent reasoning instead
uses continuous vectors in place of some intermediate
natural-language tokens.\ Let $t$ denote
the current sequence position, and let $\bm{h}_t^{(L)}$ be its
hidden state after the final Transformer layer.\ A mapping $\phi$
produces $
    \bm{z}_{t+1}
    =
    \phi(\bm{h}_t^{(L)})$, where $\bm{z}_{t+1}$ is used as the input embedding at position
$t+1$.\ The model can therefore continue reasoning without
decoding an intermediate natural language token and embedding
it again.

LatentMAS extends this process to collaboration among
agents~\citep{zou2026latentmas}.\ Consider a sequence of agents
$\mathcal{A}=\{A_1,\ldots,A_R\}$ with corresponding roles
$\mathcal{R}=\{\rho_1,\ldots,\rho_R\}$.\ Each agent $A_r$ follows
the instructions for its role $\rho_r$; for example, agents
may serve as a planner, critic, refiner, or final-answer agent.
Given a query $q$, each agent performs multiple latent
reasoning steps.\ Let
$
    \mathcal{C}_r
    =
    \bigl\{
        \bm{K}_r^{(\ell)},
        \bm{V}_r^{(\ell)}
    \bigr\}_{\ell=1}^{L}
$
denote the layer-wise KV cache available after agent $A_r$
finishes.\ This cache contains the latent context passed from
earlier agents together with the states added by $A_r$.
Passing $\mathcal{C}_r$ to the next agent lets it continue
from that context without decoding and re-encoding an
intermediate message.\ Only the final agent generates a
natural language answer.\ We use this latent communication
pipeline as our base multi-agent system.

\section{Method}
\label{sec:method}

\begin{figure*}[t]
    \centering
    \includegraphics[width=\textwidth]{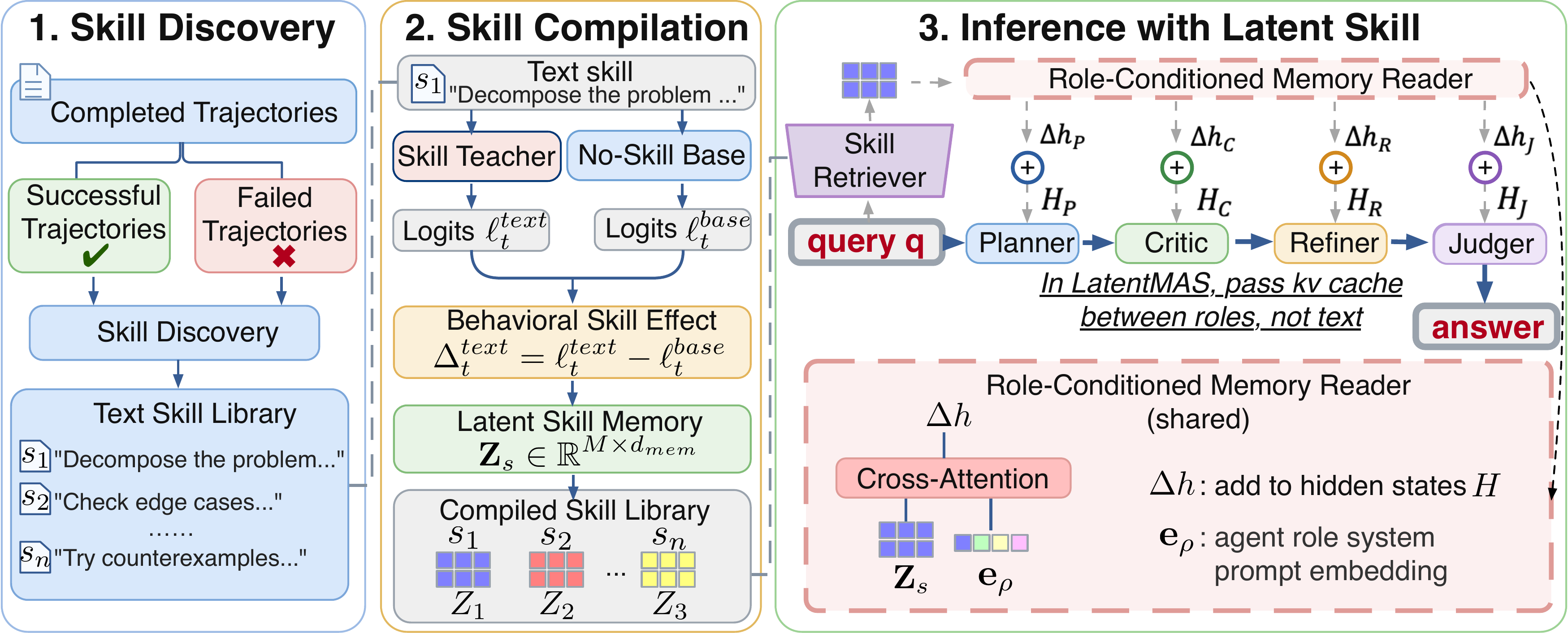}
    \caption{
    Overview of \textbf{SkillMem}.\ Reusable textual skills are discovered from paired multi-agent trajectories.\ Each skill is compiled into a compact latent
memory by matching the behavioral effect of skill injection.
At inference time, the retrieved memory is shared across agents and consumed
through a role-conditioned reader while inter-agent communication remains
entirely latent.
    }
    \label{fig:overview}
\end{figure*}

SkillMem stores each textual skill $s$ as a compact memory
$\bm{Z}_s$ that can be retrieved independently
(Section~\ref{sec:method-inject}).\ Given a task query $q$,
the system retrieves a skill $s$ and makes its memory $\bm{Z}_s$ available
to every agent in the latent MAS pipeline.\ Agents continue
to communicate through KV states, while a role-conditioned reader
uses $\bm{Z}_{s}$ to update their hidden states according to their
roles (Section~\ref{sec:method-reader}).\ We train the memories
and reader to reproduce how textual skills change the frozen
backbone's predictions (Section~\ref{sec:method-distill}).\
Skills can be provided directly or discovered from past
MAS trajectories; we use discovered skills in our
experiments (Section~\ref{sec:method-discovery}).\
Figure~\ref{fig:overview} shows the full pipeline.

\subsection{Skill Memory and Skill Retrieval}
\label{sec:method-inject}

We define a \textit{skill} as a reusable reasoning procedure that can
guide agents on multiple problem instances.\ Let
$\mathcal{S}=\{s_1,\ldots,s_N\}$ be a skill library, where
$s_i=(n_i,d_i,p_i)$ contains a name, a description of when the skill
is useful, and procedural instructions.\ Let $E(\cdot)$ denote a
text encoder based on the frozen backbone.\ We index each skill
using $\bm{e}_i=E([n_i;d_i;p_i])$.\ Given a task query $q$, we
compute $\bm{e}_q=E(q)$ and retrieve
$s_q=\arg\max_{s_i\in\mathcal{S}}
\operatorname{sim}(\bm{e}_q,\bm{e}_i)$, where
$\operatorname{sim}$ is cosine similarity.\ The selected skill
is used by every agent in the multi-agent run.

We store each skill $s$ as a persistent external memory
$\bm{Z}_s\in\mathbb{R}^{M\times d_{\mathrm{mem}}}$.\ This memory
contains $M$ learned vectors for that skill, each of dimension
$d_{\mathrm{mem}}$.\ After selecting $s$, the system
loads $\bm{Z}_{s}$ for use across agent roles.\ Each skill
has its own memory, requiring
$\mathcal{O}(M d_{\mathrm{mem}})$ skill-specific storage.

Inserting a textual skill into an agent's prompt requires
processing its tokens when the skill is used.\ Caching the
skill's layer-wise KV states avoids repeated
processing, but stores states for every skill token at every
Transformer layer.\ For a skill of $T_s$ tokens, KV storage
scales as $\mathcal{O}(L T_s d_{\mathrm{kv}})$, where $L$ is
the number of layers and $d_{\mathrm{kv}}$ is the KV
dimension.\ The storage of $\bm{Z}_s$ does not
depend on $T_s$.

\subsection{Role-Conditioned Memory Reader}
\label{sec:method-reader}

The retrieved memory $\bm{Z}_s$ is shared across agents, but its
effect should depend on the agent's role $\rho$ and current reasoning
state.\ We therefore insert a role-conditioned
\emph{cross-attention reader} at selected Transformer layers $\ell$.
For \textbf{each token}, the reader forms \textbf{queries from its hidden state}
and the \textbf{agent's role}, and forms \textbf{keys and values from
$\bm{Z}_s$}.\ The query is matched against the memory keys to determine which
memory vectors are relevant to the current token, and the
corresponding values provide the information read from memory.\
In addition, a role-conditioned gate $\bm{g}_\ell^\rho$
controls how much of this memory readout is added to the token's
hidden state.\ The reader
has its own trainable projections
$\bm{W}_q^{(\ell)}$, $\bm{W}_k^{(\ell)}$,
$\bm{W}_v^{(\ell)}$, $\bm{W}_o^{(\ell)}$, and
$\bm{W}_g^{(\ell)}$; the backbone's weights remain frozen.

We derive the role input from each agent's system instruction.
Let $\pi_\rho$ denote each role's system
instruction.\ For each role, we feed its system instruction $\pi_\rho$
into the frozen backbone and take the token representations
from a middle Transformer layer.\ We combine these representations
into a single role vector $\bm{e}_\rho$ using attention pooling, and $\bm{e}_\rho$ remains fixed
during memory training.

At an insertion layer $\ell$, let $\bm{H}_\ell$ contain the
agent's hidden states and define
$\bm{U}_\ell^\rho=
[\operatorname{LN}_\ell(\bm{H}_\ell);\bm{e}_\rho]$,
where we concatenate the role vector $\bm{e}_\rho$ with the
hidden state at each token position.\ $\operatorname{LN}_\ell$ denotes layer
normalization.
The reader computation is defined in Eq.~\ref{eq:1}, where
$d_r$ denotes the dimension of the reader's attention space.
\begin{align}
\label{eq:1}
\bm{Q}_\ell^\rho
&=\bm{U}_\ell^\rho\bm{W}_q^{(\ell)},
&
\bm{K}_{s,\ell}
&=\bm{Z}_s\bm{W}_k^{(\ell)},
&
\bm{V}_{s,\ell}
&=\bm{Z}_s\bm{W}_v^{(\ell)},
&
\bm{A}_\ell^\rho
&=\operatorname{softmax}
\left(
\frac{\bm{Q}_\ell^\rho\bm{K}_{s,\ell}^{\top}}
{\sqrt{d_r}}
\right),
\end{align}
Let $\bm{R}_\ell^\rho
=\bm{A}_\ell^\rho\bm{V}_{s,\ell}\bm{W}_o^{(\ell)}$
denote the projected memory readout.\ The hidden-state updates are
\begin{equation}
\begin{aligned}
\bm{g}_\ell^\rho
&=\sigma\!\left(
\bm{U}_\ell^\rho\bm{W}_g^{(\ell)}
\right),
\qquad
\Delta\bm{H}_\ell^\rho
=\bm{g}_\ell^\rho\odot\bm{R}_\ell^\rho,
\qquad
\bm{H}_\ell
\leftarrow
\bm{H}_\ell+
\Delta\bm{H}_\ell^\rho.
\end{aligned}
\end{equation}
Here, $\sigma(\cdot)$ is the sigmoid function, which maps each
element of $\bm{g}_\ell^\rho$ to $(0,1)$, and $\odot$ denotes
elementwise multiplication.\ Thus, $\bm{g}_\ell^\rho$ controls
the contribution of each element in the memory readout
$\bm{R}_\ell^\rho$ to the hidden-state update
$\Delta\bm{H}_\ell^\rho$.\ Each insertion layer has its own
reader parameters, shared across all skills and agent roles at
that layer.\ Therefore, each skill requires only one stored memory
$\bm{Z}_s$.\ The receiving agent's hidden states and role determine
which parts of that memory are read and how strongly they update
the hidden states.\ Agents continue to communicate through latent
states.

\subsection{Behavioral Skill Distillation}
\label{sec:method-distill}

\begin{figure*}[t]
    \centering
    \includegraphics[width=\textwidth]{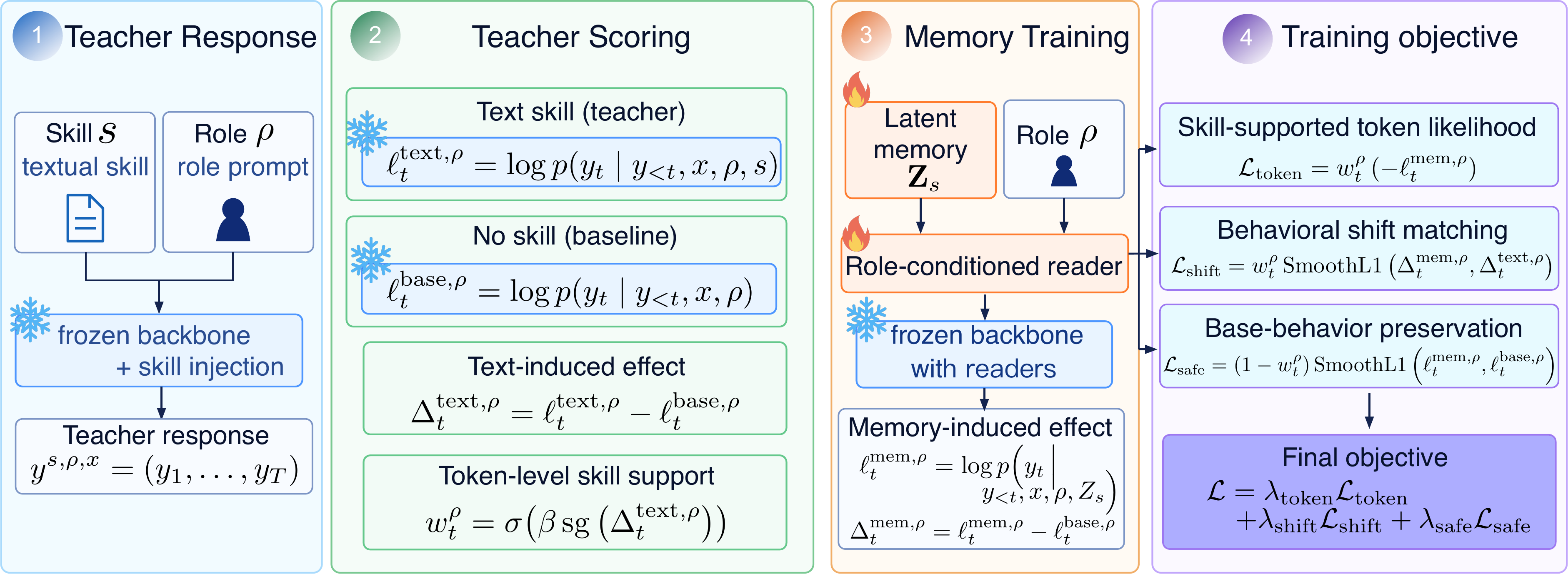}
    \caption{
    Behavioral skill distillation.\ For each skill, agent role, and
    calibration task, the frozen backbone generates a teacher response
    with the textual skill included in the role prompt.\ We score the
    same response with and without the textual skill to measure its
    token-level effect.\ We then replace the skill text with its memory
    and train the memory and role-conditioned reader to reproduce the
    skill-supported behavior while preserving the base behavior at
    positions with limited skill support.
    }
    \label{fig:distillation}
\end{figure*}

A compact memory cannot be expected to reproduce the wording of
a textual skill.\ Instead, we train it to reproduce the effect of
that skill on the model's predictions.\ For each skill, role, and
calibration task, we first generate a fixed response using the
textual skill.\ We then score the same response under three
conditions: with the textual skill, without the skill, and with
the skill memory.\ The change caused by the textual skill provides
the training target for the memory.\ During this process, we train
the skill memories and reader while keeping the backbone frozen.
Figure~\ref{fig:distillation} illustrates this procedure.

\paragraph{Teacher responses and behavioral targets.}
Consider a textual skill $s$, an agent role $\rho$, and a
calibration task $x$.\ We add the textual skill to the prompt for
role $\rho$ and use the frozen backbone to generate a teacher
response
$y^{s,\rho,x}=(y_1,\ldots,y_T)$, where $T$ is the response
length and $t\in\{1,\ldots,T\}$ indexes a token position.\ For readability, we omit the task index $x$
from the notation below.

Intermediate agents normally communicate through latent states and
do not generate text.\ For these roles, we generate a textual
response only to construct a training target.\ The generation starts
from the role prompt without the KV states passed from preceding
agents.\ These teacher responses are used only during training and
do not change the latent communication process at inference time.

We score each teacher response under three conditions.\ The
\emph{text} condition includes the textual skill in the prompt.
The \emph{base} condition uses the same prompt without the skill.
The \emph{memory} condition also removes the skill text but provides
$\bm{Z}_s$ through the role-conditioned reader.\ All three conditions
use teacher forcing: when scoring token $y_t$, the model receives
the same preceding teacher tokens $y_{<t}$.\ This ensures that the
three scores differ only in how the skill is provided.

Let $\ell_t^{\mathrm{text},\rho}$,
$\ell_t^{\mathrm{base},\rho}$, and
$\ell_t^{\mathrm{mem},\rho}$ denote the log-probability assigned
to $y_t$ under the three conditions.\ We define the text-induced
and memory-induced changes relative to the base model as
$
\Delta_t^{\mathrm{text},\rho}
=
\ell_t^{\mathrm{text},\rho}
-
\ell_t^{\mathrm{base},\rho},
\Delta_t^{\mathrm{mem},\rho}
=
\ell_t^{\mathrm{mem},\rho}
-
\ell_t^{\mathrm{base},\rho}.
$
A positive $\Delta_t^{\mathrm{text},\rho}$ means that the textual
skill increases the model's probability of the teacher token
$y_t$.\ We train the memory so that
$\Delta_t^{\mathrm{mem},\rho}$ follows
$\Delta_t^{\mathrm{text},\rho}$.

\paragraph{Token-level skill support.}
A teacher response contains both skill-related reasoning and
generic tokens that the model could predict without the skill.\
We therefore give more training weight to positions where the
textual skill increases the probability of the teacher token,
and preserve the base model's behavior where the skill provides
little support.\ Specifically, we convert the text-induced change
at each position into a soft weight:
$
w_t^\rho
=
\sigma\!\left(
\beta\,
\operatorname{sg}
\bigl(\Delta_t^{\mathrm{text},\rho}\bigr)
\right).
$
Here, $\sigma(\cdot)$ is the sigmoid function, $\beta$ controls
the sharpness of the weighting, and $\operatorname{sg}(\cdot)$
denotes stop-gradient.\ A larger $w_t^\rho$ indicates that the
textual skill provides stronger support for token $y_t$.
We use $w_t^\rho$ to emphasize skill-supported behavior and
$1-w_t^\rho$ to preserve the base behavior at positions with
limited skill support.

\paragraph{Training objective.}
The token weight $w_t^\rho$ acts as a soft switch between
imitating the textual skill and preserving the base behavior.
When $w_t^\rho$ is large, the memory-conditioned model should
follow the text-conditioned model; when it is small, the memory
should have little effect.\ We optimize
\begin{align}
\label{eq:token}
\mathcal{L}_{\mathrm{token}}
&=
\mathbb{E}_{s,\rho,x,t}
\left[
w_t^\rho
\bigl(-\ell_t^{\mathrm{mem},\rho}\bigr)
\right],
\\
\label{eq:shift}
\mathcal{L}_{\mathrm{shift}}
&=
\mathbb{E}_{s,\rho,x,t}
\left[
w_t^\rho\,
\operatorname{SmoothL1}
\left(
\Delta_t^{\mathrm{mem},\rho},
\Delta_t^{\mathrm{text},\rho}
\right)
\right],
\\
\label{eq:safe}
\mathcal{L}_{\mathrm{safe}}
&=
\mathbb{E}_{s,\rho,x,t}
\left[
(1-w_t^\rho)\,
\operatorname{SmoothL1}
\left(
\ell_t^{\mathrm{mem},\rho},
\ell_t^{\mathrm{base},\rho}
\right)
\right].
\end{align}
The expectation is taken over skills, roles, calibration tasks,
and teacher-response tokens.
$\mathcal{L}_{\mathrm{token}}$ increases the likelihood of
skill-supported teacher tokens,
$\mathcal{L}_{\mathrm{shift}}$ matches the strength of the
memory-induced and text-induced changes, and
$\mathcal{L}_{\mathrm{safe}}$ keeps the model close to its
base behavior at positions with limited skill support.
Here, $\operatorname{SmoothL1}$ is quadratic for small errors
and linear for larger errors.

Because both shifts use the same base condition,
$\Delta_t^{\mathrm{mem},\rho}
-\Delta_t^{\mathrm{text},\rho}
=
\ell_t^{\mathrm{mem},\rho}
-\ell_t^{\mathrm{text},\rho}$.
Thus, the shift loss is algebraically equivalent to matching
the memory-conditioned and text-conditioned log-probabilities.
We retain the shift notation to express both effects relative
to the no-skill behavior; the same base condition is also used
to compute $w_t^\rho$ and $\mathcal{L}_{\mathrm{safe}}$. The full objective is
$
\mathcal{L}
=
\lambda_{\mathrm{token}}\mathcal{L}_{\mathrm{token}}
+
\lambda_{\mathrm{shift}}\mathcal{L}_{\mathrm{shift}}
+
\lambda_{\mathrm{safe}}\mathcal{L}_{\mathrm{safe}}
$.
Teacher responses from all roles supervise the same
$\bm{Z}_s$; the skill memories and shared reader are trained
jointly while the backbone remains frozen.

\subsection{Skill Discovery}
\label{sec:method-discovery}

The preceding method can compile any textual skill into a latent
memory.\ To construct the skill libraries used in our experiments,
we discover such skills using the same backbone model that is
subsequently evaluated.\ For each benchmark $\mathcal{D}$, we
reserve a calibration set $\mathcal{D}_{\mathrm{cal}}$ of 20
examples.\ We run the backbone in the base multi-agent pipeline
multiple times on each calibration task.\ A completed
\emph{multi-agent trajectory} contains the successive outputs
produced by the planner, critic, refiner, and judger.\ For each
task instance $x_i$, we retain a matched pair
$\mathcal{P}_i=(\tau_i^+,\tau_i^-)$, where the final answer of
$\tau_i^+$ is correct and that of $\tau_i^-$ is incorrect.\ The
resulting discovery set is
$\mathcal{P}
=\{(\tau_i^+,\tau_i^-)\}_{i=1}^{|\mathcal{D}_{\mathrm{cal}}|}$.\ We then prompt the same backbone to compare the two trajectories
in each pair and extract a reusable procedure.\ Because the paired
trajectories solve the same task, their differences can reveal
reasoning patterns that contribute to success or failure.\ Their
role-specific outputs further indicate whether a pattern concerns
planning, critique, refinement, or final verification.\ We remove candidates that contain
instance-specific entities, values, or solution details, and
deduplicate the remaining candidates by name.\ This produces a
skill library $\mathcal{S}=\{s_1,\ldots,s_N\}$ for each
benchmark--backbone setting.\ All calibration trajectories are
disjoint from the evaluation examples.\ Further details
and per-benchmark skill counts are provided in
Appendix~\ref{app:skill-discovery}, and representative skills are
listed in Appendix~\ref{app:skill-list}.

Skill discovery provides the textual skills used in our experiments,
but it is not required by the compilation method.\ If an existing
textual skill is already available, it can be added to the library
and compiled directly into $\bm{Z}_s$ using the
procedure in Section~\ref{sec:method-distill}.\ We also explored directly deriving latent skill memories from successful and
failed trajectories without textual discovery or distillation; these zero-training
variants were substantially less reliable (see Appendix~\ref{sec:appendix-latent-discovery}).

\section{Why Text-Supervised Skills Transfer to Latent Reasoning}
\label{sec:text_latent_transfer}

SkillMem is trained from textual teacher responses but used during latent reasoning.\ The key observation is that it does not reconstruct the
skill text itself; instead, it learns the \emph{behavioral effect} induced by
the skill on the frozen backbone and applies that effect through hidden-state
readers.

\noindent\textbf{Behavioral effect as a likelihood ratio.}
Let $\bm{c}_t^\rho=(x,\rho,y_{<t})$ denote the prediction context.\ Recall that
$\Delta_t^{\mathrm{text},\rho}
=
\ell_t^{\mathrm{text},\rho}
-
\ell_t^{\mathrm{base},\rho}$.
Equivalently,
\[
\Delta_t^{\mathrm{text},\rho}
=
\log
\frac{
p_\theta(y_t\mid \bm{c}_t^\rho,s)
}{
p_\theta(y_t\mid \bm{c}_t^\rho)
},
\qquad
\Delta_t^{\mathrm{mem},\rho}
=
\log
\frac{
p_{\theta,\psi}(y_t\mid \bm{c}_t^\rho,\bm{Z}_s)
}{
p_\theta(y_t\mid \bm{c}_t^\rho)
}.
\]
Thus, behavioral distillation matches the change in model behavior caused by
the textual skill rather than its textual representation.

\noindent\textbf{Aligned latent inputs.}
LatentMAS maps a final-layer latent state back toward the model input space as
$\tilde{\bm e}_{t+1}=\bm h_t^{(L)}\bm W_a$, with
$\bm W_a\approx\bm W_{\mathrm{out}}^\dagger\bm W_{\mathrm{in}}$,
where $\bm W_{\mathrm{in}}$ and $\bm W_{\mathrm{out}}$ are the input and output
embedding matrices~\citep{zou2026latentmas}.
Let $\Phi_\ell$ denote the frozen mapping from the model input to a SkillMem
insertion layer $\ell$, and define
$\bm u_{\mathrm{text}}=\Phi_\ell(\bm e)$ and
$\bm u_{\mathrm{lat}}=\Phi_\ell(\tilde{\bm e})$.
If $\Phi_\ell$ is locally $L_\Phi$-Lipschitz, then
$\|\bm u_{\mathrm{lat}}-\bm u_{\mathrm{text}}\|
\le
L_\Phi\|\tilde{\bm e}-\bm e\|$.
Hence, the LatentMAS input-output alignment also controls the hidden-state
mismatch seen by the SkillMem reader.

\noindent\textbf{Proposition 1 (Text-to-latent behavioral transfer).}
\begin{theorybox}
Let $F_s^\rho(\bm u)$ and $G_s^\rho(\bm u)$ denote the behavioral effects
induced by the textual skill and SkillMem, respectively, at reader state
$\bm u$.\ Assume both are locally Lipschitz with constants $L_F$ and $L_G$, and
that behavioral distillation satisfies
$|G_s^\rho(\bm u_{\mathrm{text}})
-
F_s^\rho(\bm u_{\mathrm{text}})|
\le \delta$.
If
$\|\tilde{\bm e}-\bm e\|\le\epsilon_{\mathrm{align}}$, then
\[
\left|
G_s^\rho(\bm u_{\mathrm{lat}})
-
F_s^\rho(\bm u_{\mathrm{lat}})
\right|
\le
\delta
+
(L_F+L_G)L_\Phi\epsilon_{\mathrm{align}}.
\]
\end{theorybox}

The bound separates the transfer error into the behavioral distillation error
$\delta$ and the representation gap introduced by latent recurrence.
LatentMAS reduces the latter through input-output alignment before the hidden
states reach the SkillMem reader.\ SkillMem therefore transfers the
\emph{behavioral effect} of a textual skill through a shared hidden-state
interface rather than reproducing the skill text itself.\ A full derivation is
provided in Appendix~\ref{app:text_latent_transfer}.

\section{Experiments}
\label{sec:experiments}

\subsection{Experimental Setup}
\label{sec:experiments-setting}

\noindent\textbf{Backbones and benchmarks.}
We evaluate \textsc{SkillMem} using Qwen3-4B, Qwen3-8B, and
Qwen3-14B~\citep{qwen2025qwen3} as frozen backbones.\ The evaluation
includes GSM8K~\citep{cobbe2021gsm8k}, ARC-Easy and
ARC-Challenge~\citep{clark2018arc},
MBPP+~\citep{austin2021mbpp,liu2023evalplus},
MedQA~\citep{jin2021medqa}, and
GPQA-Diamond~\citep{rein2023gpqa}, covering mathematical reasoning,
QA, medical reasoning, and code
generation.\ Calibration examples used for skill discovery and
distillation are disjoint from the evaluation examples.\ Unless
otherwise stated, \textbf{results are averaged over three evaluation seeds}.\
The full evaluation protocol is provided in
Appendix~\ref{app:eval-protocol}.


\noindent\textbf{SkillMem configuration.}
Each skill memory contains $M=8$ vectors of dimension
$d_{\mathrm{mem}}=256$.\ Memory readers are inserted at layers
$\{L/4,L/2,3L/4,L-1\}$, and each agent performs eight latent
reasoning steps.\ We train the skill memories and shared reader
for 200 steps using AdamW, with $\beta=5.0$ and
$(\lambda_{\mathrm{token}},\lambda_{\mathrm{shift}},
\lambda_{\mathrm{safe}})=(1.0,0.5,0.1)$.\ 
Details are provided in Appendix~\ref{app:impl-details}.

\noindent\textbf{Baselines.}
\textbf{TextMAS} performs text-based multi-agent reasoning without
skills, while \textbf{LatentMAS} is the corresponding latent
multi-agent baseline.\ \textbf{KV Skill} represents each retrieved
skill using its cached layer-wise KV states.\ Soft Prompt (\textbf{SP})~\citep{lester2021power,li2021prefix}
learns skill-specific continuous prompt vectors and inserts them
at the input layer.\ It uses the same behavioral distillation
objective as \textsc{SkillMem}, but does not use the memory reader.\
Skill-to-LoRA (\textbf{S2L})~\citep{zhang2026skilltolora} stores
each skill in a skill-specific LoRA adapter.\ \textbf{SkillMem}
stores each skill as a compact external memory accessed through
the shared role-conditioned reader.\ All skill-based methods use
the same skill library and retrieved skill for each evaluation
example.\ S2L settings are provided in
Appendix~\ref{app:impl-details}.

\vspace{-.2cm}
\subsection{Main Results}
\label{sec:experiments-main}

Table~\ref{tab:main_results} compares \textsc{SkillMem} with
text-based and latent multi-agent baselines, as well as KV,
Soft Prompt, and LoRA skill representations, across three Qwen3
model scales.

\begin{table*}[t]
\centering
\caption{
\textbf{Main results across model scales.}
We report task accuracy (\%, ``Acc.''), average output tokens
(``Token''), and inference time in seconds (``Speed'').
``Text'' and ``Latent'' denote TextMAS and LatentMAS, respectively,
and ``SP'' denotes Soft Prompt.\ SkillMem accuracy is bolded when it exceeds either TextMAS or LatentMAS.
}
\label{tab:main_results}

\small
\setlength{\tabcolsep}{0.75pt}
\renewcommand{\arraystretch}{0.85}

\resizebox{\textwidth}{!}{
\begin{tabular}{
    l
    r
    c c c c c >{\columncolor{skillgreen}}c
    c c c c c >{\columncolor{skillgreen}}c
    c c c c c >{\columncolor{skillgreen}}c
}
\toprule

\multirow{2}{*}{\textbf{Tasks}} &
\multirow{2}{*}{\textbf{Metrics}} &
\multicolumn{6}{c}{\textbf{Qwen3-4B}} &
\multicolumn{6}{c}{\textbf{Qwen3-8B}} &
\multicolumn{6}{c}{\textbf{Qwen3-14B}}
\\

\cmidrule(lr){3-8}
\cmidrule(lr){9-14}
\cmidrule(lr){15-20}

&
&
\textbf{Text} &
\textbf{Latent} &
\textbf{KV} &
\textbf{SP} &
\textbf{S2L} &
\textbf{SkillMem}
&
\textbf{Text} &
\textbf{Latent} &
\textbf{KV} &
\textbf{SP} &
\textbf{S2L} &
\textbf{SkillMem}
&
\textbf{Text} &
\textbf{Latent} &
\textbf{KV} &
\textbf{SP} &
\textbf{S2L} &
\textbf{SkillMem}
\\

\midrule

\multirow{3}{*}{\textbf{GSM8K}}
& Acc.\ $\uparrow$
& 89.40 & 72.40 & 84.90 & 89.20 & 84.13 & \textbf{89.90}
& 90.80 & 92.33 & 92.60 & 92.20 & 92.27 & \textbf{93.07}
& 94.00 & 94.30 & 94.30 & 93.70 & 94.00 & 92.60
\\
& Token $\downarrow$
& 229 & 215 & 205 & 208 & 185 & 169
& 240 & 196 & 206 & 195 & 223 & 202
& 198 & 188 & 195 & 195 & 187 & 175
\\
& Speed $\downarrow$
& 19.1 & 6.8 & 6.2 & 6.1 & 7.6 & 5.5
& 35.1 & 6.0 & 12.2 & 6.0 & 7.9 & 6.6
& 26.8 & 9.4 & 11.8 & 9.7 & 9.7 & 8.9
\\

\midrule

\multirow{3}{*}{\textbf{GPQA-D}}
& Acc.\ $\uparrow$
& 29.94 & 17.97 & 22.46 & 30.24 & 27.54 & \textbf{35.63}
& 33.53 & 36.53 & 40.72 & 35.33 & 34.33 & \textbf{37.72}
& 37.72 & 31.74 & 28.14 & 35.33 & 34.13 & {35.33}
\\
& Token $\downarrow$
& 488 & 222 & 268 & 380 & 366 & 420
& 495 & 398 & 408 & 427 & 407 & 374
& 501 & 445 & 456 & 365 & 428 & 476
\\
& Speed $\downarrow$
& 54.5 & 9.0 & 9.6 & 10.3 & 12.6 & 12.1
& 49.7 & 15.2 & 22.7 & 16.2 & 14.1 & 13.5
& 64.9 & 21.2 & 28.4 & 18.0 & 23.1 & 22.6
\\

\midrule

\multirow{3}{*}{\textbf{ARC-E}}
& Acc.\ $\uparrow$
& 96.87 & 92.67 & 96.27 & 97.27 & 93.00 & \textbf{97.53}
& 97.60 & 98.90 & 98.40 & 98.20 & 98.53 & {97.80}
& 98.40 & 98.70 & 99.00 & 98.80 & 98.60 & {98.90}
\\
& Token $\downarrow$
& 220 & 171 & 162 & 162 & 147 & 92
& 304 & 163 & 168 & 177 & 191 & 170
& 261 & 169 & 187 & 164 & 119 & 172
\\
& Speed $\downarrow$
& 22.0 & 5.4 & 6.5 & 5.0 & 5.4 & 3.3
& 27.4 & 7.4 & 5.3 & 7.8 & 8.0 & 5.7
& 34.1 & 9.8 & 10.9 & 9.7 & 7.2 & 8.8
\\

\midrule

\multirow{3}{*}{\textbf{ARC-C}}
& Acc.\ $\uparrow$
& 90.53 & 83.67 & 86.27 & 89.80 & 82.67 & \textbf{91.20}
& 91.80 & 92.40 & 92.80 & 92.60 & 92.47 & 91.80
& 93.40 & 94.20 & 94.40 & 93.33 & 93.00 & {93.93}
\\
& Token $\downarrow$
& 294 & 190 & 186 & 177 & 193 & 162
& 341 & 183 & 191 & 205 & 195 & 163
& 315 & 189 & 210 & 153 & 137 & 222
\\
& Speed $\downarrow$
& 26.8 & 5.9 & 8.5 & 5.2 & 6.9 & 5.2
& 25.2 & 8.2 & 11.2 & 9.1 & 7.7 & 5.5
& 37.1 & 11.2 & 10.3 & 9.7 & 7.7 & 11.4
\\

\midrule

\multirow{3}{*}{\textbf{MedQA}}
& Acc.\ $\uparrow$
& 60.10 & 21.84 & 33.57 & 55.63 & 33.65 & \textbf{61.50}
& 64.56 & 67.84 & 68.55 & 66.43 & 67.14 & \textbf{69.49}
& 70.42 & 71.60 & 69.25 & 68.08 & 69.01 & 69.72
\\
& Token $\downarrow$
& 360 & 190 & 166 & 259 & 252 & 215
& 416 & 250 & 267 & 279 & 298 & 263
& 406 & 183 & 182 & 214 & 90 & 54
\\
& Speed $\downarrow$
& 44.1 & 7.7 & 11.4 & 8.3 & 10.5 & 6.8
& 49.0 & 11.5 & 8.5 & 10.6 & 10.6 & 8.6
& 61.3 & 11.7 & 10.2 & 13.3 & 6.5 & 4.8
\\

\midrule

\multirow{3}{*}{\textbf{MBPP+}}
& Acc.\ $\uparrow$
& 67.75 & 32.74 & 40.12 & 60.96 & 31.07 & {65.39}
& 65.11 & 70.23 & 70.23 & 69.65 & 68.72 & \textbf{71.56}
& 66.86 & 72.09 & 70.93 & 71.61 & 70.64 & {67.25}
\\
& Token $\downarrow$
& 187 & 122 & 121 & 96 & 163 & 78
& 92 & 76 & 79 & 78 & 77 & 68
& 121 & 87 & 109 & 102 & 118 & 129
\\
& Speed $\downarrow$
& 36.5 & 9.0 & 8.3 & 6.2 & 11.3 & 5.6
& 19.9 & 3.1 & 3.2 & 3.2 & 3.7 & 3.1
& 37.2 & 4.6 & 8.4 & 5.1 & 6.1 & 8.6
\\

\bottomrule
\end{tabular}
}
\end{table*}

\noindent\textbf{Overall performance.}
With Qwen3-4B, \textsc{SkillMem} outperforms \textsc{LatentMAS}
and all three alternative skill representations on every benchmark,
and achieves the highest overall accuracy on five of the six tasks.\
With Qwen3-8B, it achieves the best results on GSM8K, MedQA, and
MBPP+.\
Overall, the results show that compact external memory can preserve
useful skill behavior without storing layer-wise KV states or
skill-specific model parameters.

\noindent\textbf{Effect of model scale.}
The gains are strongest and most consistent with Qwen3-4B, become
task dependent with Qwen3-8B, and show no consistent advantage with
Qwen3-14B.\ This reduction is partly consistent with a ceiling effect:
on GSM8K and ARC, the 14B baselines already achieve accuracy above
90\%, leaving less room for additional skill conditioning.\ Overall,
SkillMem provides the clearest benefit when the backbone has more
room to improve from external procedural guidance.

\vspace{-10pt}
\begin{table*}[ht]
\centering
\caption{
(a) Skill storage scaling with bank size.
(b) Inference and training efficiency.
}
\label{tab:efficiency}

\footnotesize
\renewcommand{\arraystretch}{1.0}

\begin{minipage}[c]{0.56\textwidth}
\centering
\vspace{0pt}

\textbf{(a) Skill Storage Scaling}

\vspace{2pt}

\resizebox{\linewidth}{!}{
\begin{tabular}{lrrrr}
\toprule
\textbf{Method}
& \textbf{Per-Skill}
& $\mathbf{N=10}$
& $\mathbf{N=100}$
& $\mathbf{N=1000}$
\\
\midrule

Text Skill
& 173 B
& 1.7 KB
& 17.3 KB
& 173 KB
\\

LoRA Skill
& 30.7 MB
& 306.9 MB
& 3.07 GB
& 30.69 GB
\\

KV Skill
& 13.0 MB
& 135.9 MB
& 1.36 GB
& 13.59 GB
\\

\midrule

\rowcolor{skillgreen}
\textbf{SkillMem: Reader}
& \textit{shared}
& 34.5 MB
& 34.5 MB
& 34.5 MB
\\

\rowcolor{skillgreen}
\textbf{SkillMem: Memory}
& 9.6 KB
& 96 KB
& 0.96 MB
& 9.6 MB
\\

\rowcolor{skillgreen}
\textbf{SkillMem: Total}
& --
& \textbf{34.6 MB}
& \textbf{35.5 MB}
& \textbf{44.1 MB}
\\

\rowcolor{skillgreen}
\quad vs.\ KV
& --
& $\mathbf{3.9\times}$
& $\mathbf{38.3\times}$
& $\mathbf{308\times}$
\\

\rowcolor{skillgreen}
\quad vs.\ LoRA
& --
& $\mathbf{8.9\times}$
& $\mathbf{86.6\times}$
& $\mathbf{696\times}$
\\

\bottomrule
\end{tabular}
}
\end{minipage}
\hfill
\begin{minipage}[c]{0.40\textwidth}
\centering
\vspace{0pt}

\textbf{(b) Inference and Training Efficiency}

\vspace{2pt}

\resizebox{\linewidth}{!}{
\begin{tabular}{lrrr}
\toprule
\multicolumn{4}{c}{\textbf{Inference}} \\
\midrule
\textbf{Method}
& \textbf{Latency} $\downarrow$
& \textbf{Speedup} $\uparrow$
& \textbf{Mem.} $\downarrow$
\\
\midrule

TextMAS
& 33.65
& $1.00\times$
& 34.02
\\

LatentMAS
& 8.18
& $4.11\times$
& 33.83
\\

Text Skill
& 12.98
& $2.59\times$
& 36.13
\\

KV Skill
& 12.01
& $2.80\times$
& 34.97
\\

\rowcolor{skillgreen}
\textbf{SkillMem}
& \textbf{7.95}
& $\mathbf{4.23\times}$
& \textbf{33.86}
\\

\midrule
\multicolumn{4}{c}{\textbf{Training (10-skill MedQA bank)}} \\
\midrule
\textbf{Method}
& \multicolumn{2}{c}{\textbf{Scheme}}
& \textbf{Time}
\\
\midrule

LoRA Skill
& \multicolumn{2}{c}{Per-skill Adapters}
& 110 min
\\

\rowcolor{skillgreen}
\textbf{SkillMem}
& \multicolumn{2}{c}{\textbf{Shared Reader}}
& \textbf{11.7 min}
\\

\bottomrule
\end{tabular}
}

\end{minipage}

\end{table*}

\vspace{-15pt}

\subsection{Storage and Efficiency}
\label{sec:experiments-efficiency}

Table~\ref{tab:efficiency} compares the efficiency of different skill representations on Qwen3-8B.

\noindent\textbf{Storage.}
Text Skill has negligible storage but must be processed at each use.
KV Skill and LoRA Skill require 13.0\,MB and 30.7\,MB per skill,
respectively, while SkillMem uses only 9.6\,KB per skill plus one shared
reader. From 10 to 1000 skills, SkillMem is $3.9\times$--$308\times$
smaller than KV Skill.

\noindent\textbf{Inference efficiency.}
SkillMem achieves 7.95\,s end-to-end latency, comparable to LatentMAS
at 8.18\,s and lower than Text Skill and KV Skill at 12.98\,s and
12.01\,s. Its fixed eight-vector memory adds neither skill tokens nor
skill-specific KV prefixes, preserving efficient latent inference.

\noindent\textbf{Training efficiency.}
For 10 MedQA skills, SkillMem trains its shared reader in 11.7 minutes,
versus 110 minutes for separate LoRA adapters, a $9.4\times$ reduction.
This advantage comes from sharing the reader across the skill bank, whereas
S2L trains a separate adapter for each skill.

\subsection{Ablation Studies}
\label{sec:experiments-ablation}

Table~\ref{tab:ablations} evaluates the role-conditioned reader,
memory capacity, and distillation objective while keeping the
retrieved skills and evaluation examples fixed.

\noindent\textbf{Role conditioning.}
Table~\ref{tab:ablations}(a) compares the full reader with
role-agnostic, constant-embedding, shuffled-embedding, and per-role
memory variants.\ Removing role conditioning causes the largest
drop, reducing average accuracy from $75.15\%$ to $67.37\%$ on
Qwen3-4B and from $77.97\%$ to $76.50\%$ on Qwen3-8B.\ Separate
memories for each role provide no advantage over one shared memory,
supporting role-conditioned access to shared skill knowledge.

\noindent\textbf{Memory capacity.}
Table~\ref{tab:ablations}(b) shows a non-monotonic relationship
between memory size and accuracy.\ The default $M{=}8$ achieves
the best average accuracy of $64.08\%$, while $M{=}2$ remains close
at $63.63\%$ with one quarter as many memory vectors.\ Increasing
the capacity to $M{=}16$ reduces accuracy, indicating that only a
small memory is needed to capture useful skill behavior.

\noindent\textbf{Distillation objective.}
Removing any loss term reduces the macro-average accuracy from
$64.55\%$ to between $62.43\%$ and $62.93\%$ in
Table~\ref{tab:ablations}(c).\ The largest drop occurs without
$\mathcal{L}_{\mathrm{shift}}$, supporting direct matching of the
behavioral change induced by the textual skill.\ The full objective
performs best overall, showing that the three losses provide
complementary training signals.

\begin{table*}[t]
\vspace{-.8cm}
\centering
\caption{
\textbf{Ablation studies of \textsc{SkillMem}.}
Results report task accuracy (\%).
}
\label{tab:ablations}

\scriptsize
\renewcommand{\arraystretch}{0.95}
\setlength{\tabcolsep}{2.6pt}

\begin{minipage}[t]{0.57\textwidth}
\centering

\textbf{(a) Role-Conditioned Memory Access}

\vspace{3pt}

\resizebox{\linewidth}{!}{
\begin{tabular}{lcccccc}
\toprule
\textbf{Method}
& \textbf{GSM8K}
& \textbf{GPQA-D}
& \textbf{ARC-E}
& \textbf{ARC-C}
& \textbf{MedQA}
& \textbf{Avg.} \\
\midrule

\multicolumn{7}{c}{\textit{Qwen3-4B}} \\
\midrule

\textsc{SkillMem}
& 89.90
& 35.63
& 97.53
& 91.20
& 61.50
& \textbf{75.15} \\

Role Agnostic
& 75.70
& 32.94
& 97.87
& 89.73
& 40.61
& 67.37 \\

Constant Emb.
& 87.40
& 34.14
& 97.53
& 90.07
& 59.39
& 73.71 \\

Shuffled Emb.
& 85.00
& 34.13
& 97.33
& 87.80
& 57.75
& 72.40 \\

Per-role Memory
& 83.90
& 32.94
& 96.60
& 90.27
& 58.92
& 72.52 \\

\midrule
\multicolumn{7}{c}{\textit{Qwen3-8B}} \\
\midrule

\textsc{SkillMem}
& 93.07
& 37.72
& 97.80
& 91.80
& 69.48
& \textbf{77.97} \\

Role Agnostic
& 88.93
& 35.33
& 98.80
& 91.20
& 68.23
& 76.50 \\

Constant Emb.
& 92.27
& 40.12
& 98.40
& 92.20
& 66.04
& 77.81 \\

Shuffled Emb.
& 92.87
& 38.32
& 98.30
& 91.40
& 68.54
& 77.89 \\

Per-role Memory
& 90.87
& 35.93
& 98.40
& 91.60
& 66.98
& 76.76 \\

\bottomrule
\end{tabular}
}

\end{minipage}
\hfill
%
\begin{minipage}[t]{0.38\textwidth}
\centering

\textbf{(b) Memory Capacity}

\vspace{3pt}

\resizebox{\linewidth}{!}{
\begin{tabular}{lccccc}
\toprule
\textbf{Task--Model}
& $M{=}1$
& $M{=}2$
& $M{=}4$
& $M{=}8$
& $M{=}16$ \\
\midrule

GSM8K-4B
& 87.67
& 87.80
& 82.00
& \textbf{89.90}
& 85.87 \\

GSM8K-8B
& \textbf{93.47}
& 92.47
& 92.67
& 93.07
& 89.40 \\

GPQA-D-4B
& 34.53
& \textbf{36.13}
& 32.14
& 35.63
& 29.74 \\

GPQA-D-8B
& \textbf{38.12}
& \textbf{38.12}
& 37.52
& 37.72
& 35.33 \\

\bottomrule
\end{tabular}
}

\vspace{5pt}

\textbf{(c) Distillation Objective}

\vspace{3pt}

\resizebox{\linewidth}{!}{
\begin{tabular}{lcccc}
\toprule
\textbf{Task--Model}
& \textbf{Full}
& w/o $\mathcal{L}_{\mathrm{token}}$
& w/o $\mathcal{L}_{\mathrm{shift}}$
& w/o $\mathcal{L}_{\mathrm{safe}}$ \\
\midrule

GSM8K-4B  & \textbf{89.90} & 87.33 & 86.13 & 88.00 \\
GSM8K-8B  & \textbf{93.07} & 92.00 & 92.33 & 92.87 \\
GPQA-D-4B & \textbf{35.63} & 33.73 & 32.53 & 31.34 \\
GPQA-D-8B & 37.72 & \textbf{37.92} & 35.73 & 33.33 \\
MedQA-4B  & \textbf{61.50} & 59.31 & 60.56 & 60.56 \\
MedQA-8B  & \textbf{69.49} & 67.29 & 67.29 & 68.70 \\
\midrule
Avg.      & \textbf{64.55} & 62.93 & 62.43 & 62.47 \\

\bottomrule
\end{tabular}
}

\end{minipage}

\end{table*}

\section{Conclusion and Future Work}

We introduced \textsc{SkillMem}, which represents each reusable
skill as a compact external memory for latent multi-agent reasoning.\
A role-conditioned reader allows agents with different roles to
share the same memory, while behavioral distillation transfers the
effect of a textual skill without modifying the frozen backbone.\
Across six benchmarks and three model scales, the results show that
compact skill memories can provide useful procedural guidance with
low skill-specific storage and efficient latent inference.

Future work can extend \textsc{SkillMem} to retrieve and compose
multiple skills within the same reasoning process, including methods
for resolving conflicting guidance.\ Retrieval could also incorporate
the agent's role and current reasoning state in addition to the task
query.\ Finally, skill memories could be updated online as agents
encounter new tasks and interaction patterns.

\section*{AI Assistance Disclosure}
Large language models were used to assist with language editing and manuscript
polishing, including improvements to grammar, clarity, and readability. They
were also used to assist with implementing parts of the research code. All
model-assisted code was reviewed, tested, and verified by the authors. The
research ideas, methodology, experimental design, analysis, and scientific
conclusions were developed and validated by the authors, who take full
responsibility for the content of the paper.

\bibliography{iclr2027_conference}
\bibliographystyle{iclr2027_conference}

\appendix

\section{Skill Discovery}
\label{app:skill-discovery}

To discover reusable procedural skills for each benchmark, we first collect a
small calibration set using the backbone model under evaluation. For each
benchmark, we reserve 20 calibration examples that are disjoint from all
evaluation examples. Each example is run through the full text-based
multi-agent pipeline (Planner $\rightarrow$ Critic $\rightarrow$ Refiner
$\rightarrow$ Judger), using temperature $0.6$ and top-$p$ $0.95$.

For each calibration example $x_i$, we construct a matched pair of completed
multi-agent trajectories
\[
P_i = (\tau_i^{+}, \tau_i^{-}),
\]
where $\tau_i^{+}$ reaches the correct final answer and $\tau_i^{-}$ reaches an
incorrect answer for the same task. Both trajectories preserve the complete
sequence of role-specific reasoning stages, including planning, critique,
refinement, and final verification. This matched construction allows skill
discovery to compare successful and failed reasoning under the same task
conditions rather than aggregating unrelated success and failure examples.

We then prompt the same backbone model used to generate the trajectories to
compare each matched pair and identify reusable procedures. The comparison can
draw on outputs from any agent role, allowing the model to identify procedures
relevant to planning, critique, refinement, or verification. Each candidate
skill is represented as
\[
s_j = (n_j,d_j,p_j),
\]
where $n_j$ is a short name, $d_j$ describes when the skill is useful, and
$p_j$ specifies the procedure.

To reduce instance-specific artifacts, we remove candidates containing
entities, numerical values, or solution details tied to individual calibration
examples. We then deduplicate semantically overlapping candidates to obtain the
final skill library
\[
\mathcal{S} = \{s_1,\ldots,s_N\}.
\]
Skill discovery is performed independently for each backbone scale, so the
resulting libraries may differ across Qwen3-4B, Qwen3-8B, and Qwen3-14B.
Table~\ref{tab:skill-counts} reports the number of discovered skills for each
benchmark and model scale. The complete skill lists are provided in
Appendix~\ref{app:skill-list}.

\begin{table}[h]
\centering
\small
\caption{
Number of discovered skills per benchmark and model scale after filtering
instance-specific candidates and deduplicating overlapping skills.
}
\label{tab:skill-counts}
\begin{tabular}{lccc}
\toprule
\textbf{Benchmark} & \textbf{Qwen3-4B} & \textbf{Qwen3-8B} & \textbf{Qwen3-14B} \\
\midrule
GSM8K   & 10 & 10 & 8  \\
GPQA-D  & 9  & 8  & 8  \\
ARC-E   & 10 & 10 & 10 \\
ARC-C   & 9  & 9  & 9  \\
MedQA   & 8  & 10 & 8  \\
MBPP+   & 10 & 10 & 10 \\
\bottomrule
\end{tabular}
\end{table}

\subsection{Discovered Skills}
\label{app:skill-list}

Table~\ref{tab:skill-list} lists the skills discovered for each benchmark under the Qwen3-4B backbone
(the discovery process is run independently per model scale; skill sets for Qwen3-8B/14B are
qualitatively similar but not identical, since discovery reflects each backbone's own error and
success patterns).

\begingroup
\small
\setlength{\tabcolsep}{4pt}
\renewcommand{\arraystretch}{1.15}
\begin{longtable}{
  @{}
  >{\raggedright\arraybackslash}p{0.08\linewidth}
  >{\raggedright\arraybackslash}p{0.19\linewidth}
  >{\raggedright\arraybackslash}p{0.60\linewidth}
  @{}
}
\caption{Discovered skills per benchmark (Qwen3-4B backbone).} \label{tab:skill-list} \\
\toprule
\textbf{Dataset} & \textbf{Skill} & \textbf{Description} \\
\midrule
\endfirsthead
\multicolumn{3}{c}{\tablename\ \thetable{} -- continued} \\
\toprule
\textbf{Dataset} & \textbf{Skill} & \textbf{Description} \\
\midrule
\endhead
\bottomrule
\endfoot

\multirow{10}{*}{GSM8K}
 & Rate vs Total Confusion & When dealing with per-unit rates, ensure that the total amount is not confused with the rate itself. \\
 & Quantity Tracking & Track named quantities, updates, and dependencies in a word problem. \\
 & Equation Construction & Translate verbal relations into arithmetic equations before solving. \\
 & Rate and Proportion Reasoning & Handle per-unit rates, totals, fractions, percentages, and repeated events. \\
 & Intermediate Result Verification & Check intermediate arithmetic and whether each number answers the intended subquestion. \\
 & Target Quantity Restatement & Restate the question's target quantity before the final step to avoid answering a different quantity. \\
 & Unit Normalization Before Combining & Convert every rate to a common time unit before combining them. \\
 & Incremental Step Verification & Verify each step of the calculation by explicitly stating the intermediate result. \\
 & Relationship Clarification & Clarify relationships between quantities (e.g., ``twice as many'') to ensure correct proportional calculation. \\
 & Total Distance Calculation & Calculate total distance by summing individual segments, ensuring all parts are accounted for. \\
\midrule

\multirow{9}{*}{GPQA-D}
 & Rate vs Total Confusion & Avoid conflating per-unit rates with total amounts when a quantity depends on multiple variables. \\
 & Stale Intermediate Value Reuse & Avoid reusing an intermediate value from an earlier step when a different quantity is asked for. \\
 & Approximation as Exact & Avoid treating an approximation as exact when precise calculation is required. \\
 & Dimensional and Scale Analysis & Ensure units and scales are consistent and physical-dimension relationships are correctly interpreted. \\
 & Misapplication of Proportionality & Avoid assuming a linear relationship where a non-linear one is required. \\
 & Target Quantity Restatement & Restate the question's target quantity before the final step. \\
 & Unit Normalization Before Combining & Convert all rates/quantities to a common unit before combining. \\
 & First Principles Modeling & Translate an advanced science question into governing principles and variables. \\
 & Equation and Assumption Tracking & Track formulas, assumptions, approximations, and variable meanings. \\
\midrule

\multirow{10}{*}{ARC-E}
 & Rate vs Total Confusion & Avoid confusing a per-unit rate with a total amount when interpreting a physical process. \\
 & Stale Intermediate Value Reuse & Avoid reusing an intermediate value instead of the one the question asks for. \\
 & Conceptual Misalignment in Biological Processes & Avoid misinterpreting the biological process described (e.g., confusing adaptation with a specific mechanism). \\
 & Option Evaluation Misalignment & Avoid evaluating options against incorrect assumptions about their relevance. \\
 & Direct Result vs Indirect Consequence Confusion & Distinguish a direct result of a process from an indirect or secondary effect. \\
 & Target Quantity Restatement & Restate the question's target quantity before the final step. \\
 & Unit Normalization Before Combining & Convert every rate to a common unit before combining. \\
 & Specific Function Matching & Match the function of each option to the specific requirement of the question. \\
 & Key Concept Identification & Identify the core concept or principle the question is testing. \\
 & Sequence of Actions in Lab Procedures & Follow a specific sequence of actions in laboratory procedures. \\
\midrule

\multirow{9}{*}{ARC-C}
 & Rate vs Total Confusion & Distinguish per-unit rates (e.g., gravitational acceleration) from total quantities (e.g., potential energy). \\
 & Stale Intermediate Value Reuse & Avoid reusing intermediate values when the question asks for a different quantity or condition. \\
 & Infectious vs Non-Infectious Misclassification & Distinguish infectious from non-infectious diseases by transmission method, not symptoms alone. \\
 & Causal Mechanism Reasoning & Infer direct causes and effects in physical, biological, and earth-science settings. \\
 & Experiment Variable Analysis & Separate manipulated variables, measured variables, controls, and conclusions. \\
 & Target Quantity Restatement & Restate the question's target quantity before the final step. \\
 & Unit Normalization Before Combining & Convert all measurements to a common unit before combining. \\
 & Process Sequence Reasoning & Break a complex process into sequential, interdependent steps. \\
 & Ecological Context Analysis & Use ecological knowledge to interpret species-interaction and environmental-change questions. \\
\midrule

\multirow{8}{*}{MedQA}
 & Stale Intermediate Value Reuse & Avoid reusing an earlier intermediate value without verifying it is still relevant. \\
 & Treats Approximation as Exact & Do not assume an approximate value is exact; account for its precision. \\
 & Clinical Feature Prioritization & Extract high-yield symptoms, exam findings, labs, demographics, and timeline. \\
 & Differential Diagnosis & Compare plausible diagnoses using distinguishing clinical features. \\
 & Pathophysiology to Management & Use disease mechanism to select the appropriate test, treatment, or complication. \\
 & Target Quantity Restatement & Restate the question's target quantity before the final step. \\
 & Unit Normalization Before Combining & Convert relevant rates/measurements to a common unit before combining. \\
 & Complication Risk Assessment & Evaluate long-term complications or risks associated with a diagnosis. \\
\midrule

\multirow{10}{*}{MBPP+}
 & Rate vs Total Confusion & Check whether a per-unit rate is conflated with a total amount. \\
 & Stale Intermediate Value Reuse & Check whether an intermediate value is reused without being updated. \\
 & Sign Drop in Loss and Gain & Check whether the sign is dropped when combining a loss and a gain. \\
 & Quantity Mismatch in Question & Check whether the answer targets an earlier-computed quantity instead of the one asked for. \\
 & Approximation as Exact & Check whether an approximation is treated as exact without re-checking its scale. \\
 & Identify Correct Parameters & Ensure the function signature matches the required parameters. \\
 & Direct Formula Implementation & Implement the required formula directly without unnecessary intermediate steps. \\
 & Handle Edge Cases & Ensure the function handles edge cases such as zero or negative values. \\
 & Input Output Contract Analysis & Infer the required function behavior, input types, output type, and edge cases. \\
 & Algorithm Decomposition & Break the programming task into simple transformation steps before coding. \\
\end{longtable}
\endgroup

\section{Skill Retrieval}
\label{app:retrieval}

Each discovered skill is indexed once, offline, by embedding its skill name, description, and skill
prompt (concatenated) with the same frozen backbone used for evaluation: we run a forward pass with
\texttt{output\_hidden\_states=True}, take the \emph{last} hidden layer, apply attention-mask-weighted
mean pooling over token positions, and $L_2$-normalize the resulting vector. When source questions
from the discovery trajectories are available, we additionally embed each source question with the
same procedure and aggregate the resulting similarity scores with the description-level score by
taking their \emph{mean} (rather than using the skill description alone).

At inference time, the input question is embedded with the identical procedure, and the top-$1$ skill
is retrieved by cosine similarity between the query embedding and each skill's indexed embedding.
Retrieval is performed once per evaluation example (not per agent role): all agents in the same
multi-agent rollout for a given question consume the same retrieved skill. We do not apply a
minimum-similarity threshold for skill selection (threshold $=0$), so every routed example is assigned
a skill; we found that a small positive threshold instead routes a large majority of examples to
``no skill'' and understates the benefit of skill augmentation. Retrieval is precomputed once per
benchmark and cached, and the same cached routing decisions are reused across every skill-based method
in Table~1 (KV Skill, S2L, and SkillMem) to ensure a matched comparison.

\section{Evaluation Protocol}
\label{app:eval-protocol}

Table~\ref{tab:decoding-hparams} lists the decoding configuration used throughout evaluation. All
stochastic generation (calibration rollouts and every agent's turn at evaluation time) samples at
temperature $0.6$ with top-$p$ $0.95$; each agent performs $8$ latent reasoning steps before the
final (judger) role decodes a textual answer with a maximum of $512$ new tokens. Every reported
accuracy is the average over $3$ evaluation seeds ($41$, $42$, $43$), which vary only the decoding
RNG on top of a fixed trained checkpoint.

\begin{table}[h]
\centering
\caption{Decoding configuration used for evaluation.}
\label{tab:decoding-hparams}
\begin{tabular}{ll}
\toprule
\textbf{Setting} & \textbf{Value} \\
\midrule
Sampling temperature & 0.6 \\
Top-$p$ & 0.95 \\
Latent reasoning steps per agent & 8 \\
Max.\ new tokens (judger decode), default & 512 \\
Evaluation seeds (averaged) & 41, 42, 43 \\
\bottomrule
\end{tabular}
\end{table}

\section{Experimental Setup: Additional Implementation Details}
\label{app:impl-details}

\subsection{SkillMem Hyperparameters}
Table~\ref{tab:skillmem-hparams} lists the hyperparameters used to train SkillMem's latent memories and
shared role-conditioned reader, held fixed across all main-table experiments (Table~1) unless
otherwise noted in an ablation.

\begin{table}[h]
\centering
\caption{SkillMem training hyperparameters (default configuration, $M=8$).}
\label{tab:skillmem-hparams}
\begin{tabular}{ll}
\toprule
\textbf{Hyperparameter} & \textbf{Value} \\
\midrule
Memory slots per skill ($M$) & 8 \\
Memory dimension ($d_{\text{mem}}$) & 256 \\
Reader dimension ($d_r$) & 256 (defaults to $d_{\text{mem}}$) \\
Reader insertion layers & $\{L/4,\ L/2,\ 3L/4,\ L-1\}$ \\
\quad Qwen3-4B / 8B ($L{=}36$) & layers $\{9, 18, 27, 35\}$ \\
\quad Qwen3-14B ($L{=}40$) & layers $\{10, 20, 30, 39\}$ \\
Role embedding dimension & 64 \\
Role embedding extraction layer & middle decoder layer ($\lfloor L/2 \rfloor$) \\
Latent reasoning steps per agent & 8 \\
Optimizer & AdamW \\
Training steps & 200 \\
Sigmoid weighting sharpness $\beta$ & 5.0 \\
Loss weights $(\lambda_{\text{token}}, \lambda_{\text{shift}}, \lambda_{\text{safe}})$ & $(1.0,\ 0.5,\ 0.1)$ \\
Backbone & frozen throughout training \\
\bottomrule
\end{tabular}
\end{table}

\paragraph{Role representation.} For each agent role $\rho$, we encode its system-instruction text
$\pi_\rho$ with the frozen backbone and pool hidden states from the middle decoder layer
($\lfloor L/2 \rfloor$, i.e. layer 18 for Qwen3-4B/8B and layer 20 for Qwen3-14B) using a
\emph{fixed} (untrained) attention query and projection: both are drawn once from a seeded random
Gaussian initialization and kept frozen for the lifetime of the model wrapper, computed under
\texttt{no\_grad} and never registered as trainable parameters. The resulting role embedding
$e_\rho \in \mathbb{R}^{64}$ is therefore not learned; the four agent roles are distinguished purely by
the content of their respective instruction text, not by any trained role-specific parameters. This
keeps the shared reader's role-conditioning entirely a function of frozen backbone representations,
introducing no additional trainable role-specific parameters.

\paragraph{Reader parameters are not shared across insertion layers.} Each of the four insertion
layers has its own independent set of reader weights ($W_q, W_k, W_v, W_o, W_g$, plus a LayerNorm):
readers at different depths are separately parameterized modules, not a single reader applied
repeatedly. Only the memory bank $Z_s$ is shared as a single per-skill vector across all four readers;
each reader learns its own layer-specific mapping from that shared memory into its local residual
stream.

\subsection{Intermediate-Role Behavioral Distillation}
\label{app:intermediate-role}
Because \textsc{LatentMAS} agents other than the final (judger) role never emit text during normal
inference, obtaining a token-level supervision target for their behavioral-distillation losses
(Eqs.~\ref{eq:token}--\ref{eq:safe}) requires generating text for those roles \emph{during training only}. For every
(skill, role) pair, we generate one teacher continuation per calibration example by running the
frozen backbone's ordinary text-generation path -- with the retrieved skill's text injected into that
role's own prompt and \emph{no} context from other agents ($\text{context}=\varnothing$) -- and decode
greedily. This continuation is cached once and reused for all subsequent training steps; it is never
regenerated on the fly. The same $\text{context}=\varnothing$ convention is used at inference time (
cross-agent information is carried entirely through the shared latent KV cache, not through the
prompt text), so the prompt text seen at training time exactly matches the prompt text seen at
inference time for every role. The one difference between the two settings is that, at training time,
the teacher continuation for a given role is generated from a fresh KV state (no upstream latent
context), whereas at inference time the same role's turn is conditioned on the latent KV cache
accumulated from preceding agents; the textual content of the prompt itself does not depend on this
KV prefix. The generated teacher tokens serve solely as a fixed teacher-forcing target for computing
$l_{\text{text}}$, $l_{\text{base}}$, and (during training) $l_{\text{mem}}$: they are never fed back
into the model as generation input and never appear anywhere in the deployed (inference-time) pipeline,
which loads only the trained memory bank and reader weights.

Because the planner, critic, and refiner roles are, by design, not instructed to produce a final
answer, we do not apply an answer-correctness filter to their generated teacher continuations (doing
so would be a category error, since these roles have no notion of a ``final answer'' to check); only
the judger's continuation is filtered by answer correctness before being accepted into the teacher
cache.

\subsection{Skill-to-LoRA (S2L) Baseline Hyperparameters}
Table~\ref{tab:s2l-hparams} lists the hyperparameters used for the Skill-to-LoRA baseline
\citep{zhang2026skilltolora}, matched to the paper's reported architecture (rank, alpha, target
modules) with the training loop otherwise following this work's own conventions (AdamW, fixed step
budget) rather than the original paper's QLoRA recipe. One independent LoRA adapter is trained per
discovered skill, pooling teacher-forcing supervision from all four agent roles (the direct analogue
of the \emph{Role Agnostic} condition in Table~3(a), since the LoRA baseline has no native mechanism
for role conditioning).

\begin{table}[h]
\centering
\caption{Skill-to-LoRA (S2L) baseline hyperparameters.}
\label{tab:s2l-hparams}
\begin{tabular}{ll}
\toprule
\textbf{Hyperparameter} & \textbf{Value} \\
\midrule
LoRA rank $r$ & 16 \\
LoRA $\alpha$ & 32 \\
LoRA dropout & 0.0 \\
Target modules & $W_q$, $W_v$ \\
Learning rate & $1\mathrm{e}{-4}$ \\
Batch size & 4 \\
Training steps per skill & 200 \\
Optimizer & AdamW \\
Backbone & frozen throughout training \\
\bottomrule
\end{tabular}
\end{table}

\section{Detailed Analysis of Text-to-Latent Skill Transfer}
\label{app:text_latent_transfer}

This appendix provides the derivation behind
Proposition~1 in Section~\ref{sec:text_latent_transfer}.
The analysis separates text-to-latent transfer into two steps:
(i) LatentMAS maps a final-layer latent state back toward the model input
distribution, and
(ii) this input-space alignment propagates through the frozen backbone to the
hidden states consumed by the SkillMem reader.

\subsection{Input-Output Alignment in Latent Reasoning}
\label{app:latent_alignment}

In standard autoregressive generation, a discrete token is mapped through the
input embedding matrix $\bm W_{\mathrm{in}}$ before being processed by the
Transformer. In latent reasoning, LatentMAS instead continues from the
final-layer hidden state $\bm h_t^{(L)}$.
Because final-layer hidden states and learned input embeddings generally follow
different activation distributions, directly feeding $\bm h_t^{(L)}$ back as
the next input may introduce distribution mismatch.

LatentMAS addresses this by applying an input-output alignment operator
\citep{zou2026latentmas}:
\begin{equation}
\tilde{\bm e}_{t+1}
=
\bm h_t^{(L)}\bm W_a,
\qquad
\bm W_a
\approx
\bm W_{\mathrm{out}}^\dagger
\bm W_{\mathrm{in}},
\label{eq:app_alignment}
\end{equation}
where $\bm W_{\mathrm{out}}^\dagger$ denotes the pseudoinverse of the output
embedding matrix.
The aligned vector $\tilde{\bm e}_{t+1}$ is then used as the next latent input.

For the transfer analysis, let $\bm e$ denote a text-derived input embedding
and $\tilde{\bm e}$ an aligned latent input representing a nearby reasoning
state. We characterize the residual input mismatch by
\begin{equation}
\|\tilde{\bm e}-\bm e\|
\leq
\epsilon_{\mathrm{align}}.
\label{eq:app_alignment_error}
\end{equation}
We do not assume that the two inputs are identical. The quantity
$\epsilon_{\mathrm{align}}$ simply measures the remaining discrepancy after
the LatentMAS alignment step.

\subsection{From Input Alignment to Reader-State Alignment}
\label{app:hidden_alignment}

SkillMem does not directly read the final-layer latent state
$\bm h_t^{(L)}$. Instead, the aligned input is processed by the same frozen
backbone, and the role-conditioned reader operates on intermediate hidden
states at the selected insertion layers.

Let $\Phi_\ell$ denote the frozen mapping from the model input to insertion
layer $\ell$, conditioned on the preceding context. We define
\begin{equation}
\bm u_{\mathrm{text}}
=
\Phi_\ell(\bm e),
\qquad
\bm u_{\mathrm{lat}}
=
\Phi_\ell(\tilde{\bm e}),
\label{eq:app_reader_states}
\end{equation}
where $\bm u_{\mathrm{text}}$ and $\bm u_{\mathrm{lat}}$ are the reader inputs
induced by text-derived and latent reasoning, respectively.

Assume that $\Phi_\ell$ is locally Lipschitz over the region visited by these
states, with constant $L_\Phi$:
\begin{equation}
\|
\Phi_\ell(\bm e_1)-\Phi_\ell(\bm e_2)
\|
\leq
L_\Phi
\|\bm e_1-\bm e_2\|.
\label{eq:app_phi_lipschitz}
\end{equation}
Applying this condition to $\bm e$ and $\tilde{\bm e}$ gives
\begin{equation}
\|
\bm u_{\mathrm{lat}}
-
\bm u_{\mathrm{text}}
\|
\leq
L_\Phi
\epsilon_{\mathrm{align}}.
\label{eq:app_hidden_gap}
\end{equation}

Equation~\ref{eq:app_hidden_gap} connects the alignment mechanism of
LatentMAS to SkillMem. If the aligned latent input remains close to the
text-derived input space, then the hidden states presented to the reader also
remain close, up to the local sensitivity of the frozen backbone.

\subsection{Behavioral Effect Matching}
\label{app:behavioral_effect}

SkillMem is trained to reproduce the behavioral change induced by a textual
skill rather than its textual or hidden representation.
For prediction context
$\bm c_t^\rho=(x,\rho,y_{<t})$, the text-induced and memory-induced effects are
\begin{equation}
\Delta_t^{\mathrm{text},\rho}
=
\log
\frac{
p_\theta(y_t\mid\bm c_t^\rho,s)
}{
p_\theta(y_t\mid\bm c_t^\rho)
},
\qquad
\Delta_t^{\mathrm{mem},\rho}
=
\log
\frac{
p_{\theta,\psi}(y_t\mid\bm c_t^\rho,\bm Z_s)
}{
p_\theta(y_t\mid\bm c_t^\rho)
}.
\label{eq:app_behavior_ratio}
\end{equation}
These are exactly the changes relative to the shared no-skill condition used
in the behavioral distillation objective.

Let $F_s^\rho(\bm u)$ denote the behavioral effect induced by the textual
skill when the relevant hidden state is $\bm u$, and let
$G_s^\rho(\bm u)$ denote the corresponding effect induced by SkillMem.
Behavioral distillation aims to make these two effects close on the
text-derived training states. We write this as
\begin{equation}
\left|
G_s^\rho(\bm u_{\mathrm{text}})
-
F_s^\rho(\bm u_{\mathrm{text}})
\right|
\leq
\delta,
\label{eq:app_distillation_error}
\end{equation}
where $\delta$ denotes the residual distillation error.

We additionally assume that both behavioral-effect functions are locally
Lipschitz in the reader-state region:
\begin{align}
\left|
F_s^\rho(\bm u_1)-F_s^\rho(\bm u_2)
\right|
&\leq
L_F\|\bm u_1-\bm u_2\|,
\label{eq:app_F_lipschitz}
\\
\left|
G_s^\rho(\bm u_1)-G_s^\rho(\bm u_2)
\right|
&\leq
L_G\|\bm u_1-\bm u_2\|.
\label{eq:app_G_lipschitz}
\end{align}

\subsection{Proof of Text-to-Latent Behavioral Transfer}
\label{app:transfer_proof}

We now bound the difference between the memory-induced and text-induced effects
at a latent reasoning state. By adding and subtracting the corresponding
effects at the text-derived state,
\begin{align}
&
\left|
G_s^\rho(\bm u_{\mathrm{lat}})
-
F_s^\rho(\bm u_{\mathrm{lat}})
\right|
\nonumber\\
\leq\;&
\left|
G_s^\rho(\bm u_{\mathrm{lat}})
-
G_s^\rho(\bm u_{\mathrm{text}})
\right|
\nonumber\\
&+
\left|
G_s^\rho(\bm u_{\mathrm{text}})
-
F_s^\rho(\bm u_{\mathrm{text}})
\right|
\nonumber\\
&+
\left|
F_s^\rho(\bm u_{\mathrm{text}})
-
F_s^\rho(\bm u_{\mathrm{lat}})
\right|.
\label{eq:app_triangle}
\end{align}
Using Equations~\ref{eq:app_distillation_error},
\ref{eq:app_F_lipschitz}, and~\ref{eq:app_G_lipschitz}, we obtain
\begin{equation}
\left|
G_s^\rho(\bm u_{\mathrm{lat}})
-
F_s^\rho(\bm u_{\mathrm{lat}})
\right|
\leq
\delta
+
(L_F+L_G)
\|
\bm u_{\mathrm{lat}}
-
\bm u_{\mathrm{text}}
\|.
\label{eq:app_hidden_transfer}
\end{equation}
Finally, substituting the hidden-state alignment bound from
Equation~\ref{eq:app_hidden_gap} yields
\begin{equation}
\boxed{
\left|
G_s^\rho(\bm u_{\mathrm{lat}})
-
F_s^\rho(\bm u_{\mathrm{lat}})
\right|
\leq
\delta
+
(L_F+L_G)L_\Phi
\epsilon_{\mathrm{align}}
}.
\label{eq:app_final_bound}
\end{equation}

This establishes Proposition~1.

\paragraph{Interpretation.}
The bound contains two terms with different meanings.
The first, $\delta$, measures how accurately behavioral distillation reproduces
the textual skill effect on the states observed during training.
The second,
$(L_F+L_G)L_\Phi\epsilon_{\mathrm{align}}$,
captures the additional error introduced when the learned skill is applied to
latent reasoning states.
LatentMAS directly targets the input component of this term through its
input-output alignment operator, while the frozen backbone maps both text-derived
and aligned latent inputs into the same intermediate hidden-state interface.

The analysis therefore does not require textual and latent reasoning states to
be identical.
It only states that behavioral transfer remains controlled when the aligned
latent trajectory stays within a local region where the frozen backbone and
the two skill-effect functions vary smoothly.

\section{Directly Discover Latent Vector as Skills}
\label{sec:appendix-latent-discovery}

The pipeline described in the main text discovers skills in two stages:
the backbone model compares calibration trajectories and produces a
natural-language skill description, after which behavioral distillation
compiles that description into the memory vector $\mathbf{Z}_s$ used at inference. A natural question is whether both stages can be
skipped, discovering $Z$ directly from the base MAS model's own hidden states, with zero gradient updates. We
explored two variants of this idea.

\paragraph{Contrastive extraction.} For each calibration question we sample 8 attempts under temperature
decoding with the TextMAS carrier and keep any pair where at least one attempt reached the correct final
answer and at least one did not (MedQA and GPQA-D yielded 13 such pairs out of 20 calibration questions each;
GSM8K yielded only 5, since its baseline accuracy is high enough that few sampled attempts fail). For each pair
we render the two trajectories to text and extract a representation from \emph{each} side using the same
mechanism already used to initialize $Z$ from skill text: hidden states from a fixed decoder layer are pooled
by attention against $M{=}8$ fixed random queries and projected to $d_{\text{mem}}{=}256$ by a fixed random
matrix. The skill vector for a pair is the difference $Z_{\text{pair}} = Z_{\text{success}} - Z_{\text{failure}}$,
and the final $Z_{\text{skill}}$ is the mean of $Z_{\text{pair}}$ over all valid pairs. The raw extracted norm
is roughly two orders of magnitude smaller than a gradient-trained $Z$ (e.g.\ $\sim$1--11 vs.\ $\sim$335--360),
so we rescale $Z_{\text{skill}}$ to match the mean norm of an already-trained checkpoint's skills before
evaluating it through that checkpoint's (already-trained, skill-agnostic) cross-attention reader.

\paragraph{Latent generation.} Rather than hand-extracting a direction, this variant lets the base model
generate one. We concatenate several (success, failure) pairs into a single reflection prompt instructing the
model to identify what separates the correct attempts from the incorrect ones, then run $M{=}8$ steps of the
model's own continuous latent reasoning -- the identical mechanism the Planner/Critic/Refiner agents already
use for their internal thinking, with no text ever decoded. The resulting $M$ "thought" embeddings are
projected to $d_{\text{mem}}$ with the same fixed random map and rescaled identically to the extraction variant.

\paragraph{Results.} Both variants require zero gradient training and reuse an already-trained, shared
cross-attention reader; only $Z$ itself is produced without training. Table~\ref{tab:latent-discovery} reports
the 3-seed accuracy delta against the no-skill LatentMAS baseline on Qwen3-8B, forcing the discovered skill on
every held-out example (no routing).

\begin{table}[h]
\centering
\begin{tabular}{lrr}
\toprule
Benchmark & Contrastive extraction $\Delta$ & Latent generation $\Delta$ \\
\midrule
MedQA   & $+0.00$ pp  & $-1.57$ pp \\
GSM8K   & $+0.27$ pp  & $-0.13$ pp \\
GPQA-D  & $-32.53$ pp & $-4.96$ pp \\
\bottomrule
\end{tabular}
\caption{3-seed accuracy delta vs.\ no-skill LatentMAS for the two zero-training, latent-space skill
discovery variants (Qwen3-8B).}
\label{tab:latent-discovery}
\end{table}

Contrastive extraction matches or marginally beats the no-skill baseline on MedQA and GSM8K, but on GPQA-D the
extracted direction destabilizes generation outright: outputs collapse into degenerate repetition (e.g.\ an
unbounded repeated fragment) rather than producing a final answer, consistently across all three seeds.
Inspecting the pairs' internal consistency does not explain this in advance -- GSM8K's pairs are the most
mutually consistent of the three benchmarks yet produce only a noise-level effect, while MedQA and GPQA-D have
similar consistency but opposite outcomes. Latent generation never collapses on any benchmark (verified by
inspecting raw predictions before committing to a full run) but also never exceeds the no-skill baseline.

Neither variant is competitive with the text-skill-plus-distillation pipeline used elsewhere in this work.
The comparison is nonetheless informative: it indicates that the discovery and distillation stages are not
easily bypassed by directly manipulating hidden states with an off-the-shelf, untrained reader -- the reader's
cross-attention weights are shaped by gradient training on a particular distribution of $Z$, and an
out-of-distribution direction can be safe on some benchmarks and destructive on others, with no obvious way to
predict which in advance.

\end{document}